\documentclass{article}

\usepackage{PRIMEarxiv}

\usepackage[utf8]{inputenc} 
\usepackage[T1]{fontenc}    
\usepackage{hyperref}       
\usepackage{url}            
\usepackage{booktabs}       
\usepackage{amsfonts}       
\usepackage{nicefrac}       
\usepackage{microtype}      
\usepackage{lipsum}
\usepackage{fancyhdr}       
\usepackage{graphicx}       
\usepackage{natbib}
\usepackage{makecell}
\usepackage{subfigure}
\graphicspath{{media/}}     

\title{Fast Polypharmacy Side Effect Prediction Using Tensor Factorisation
}

\author{
  Oliver Lloyd\\
  MRC Integrative Epidemiology Unit \\
  University of Bristol \\
  Bristol, UK\\
  \texttt{oliver.lloyd@bristol.ac.uk} \\
   \And
  Yi Liu, Tom R. Gaunt \\
  MRC Integrative Epidemiology Unit \\
  University of Bristol \\
  Bristol, UK\\
}

\begin{document}
\maketitle

\begin{abstract}
\textbf{Motivation:} Adverse reactions from drug combinations are increasingly common, making their accurate prediction a crucial challenge in modern medicine. Laboratory-based identification of these reactions is insufficient due to the combinatorial nature of the problem. While many computational approaches have been proposed, tensor factorisation models have shown mixed results, necessitating a thorough investigation of their capabilities when properly optimized. \textbf{Results:} We demonstrate that tensor factorisation models can achieve state-of-the-art performance on polypharmacy side effect prediction, with our best model (SimplE) achieving median scores of 0.978 AUROC, 0.971 AUPRC, and 1.000 AP@50 across 963 side effects. Notably, this model reaches 98.3\% of its maximum performance after just two epochs of training (approximately 4 minutes), making it substantially faster than existing approaches while maintaining comparable accuracy. We also find that incorporating monopharmacy data as self-looping edges in the graph performs marginally better than using it to initialize embeddings. \textbf{Availability and Implementation:} All code used in the experiments is available in our GitHub repository (\href{https://doi.org/10.5281/zenodo.10684402}{https://doi.org/10.5281/zenodo.10684402}). The implementation was carried out using Python 3.8.12 with PyTorch 1.7.1, accelerated with CUDA 11.4 on NVIDIA GeForce RTX 2080 Ti GPUs. \textbf{Contact:} \href{email:oliver.lloyd@bristol.ac.uk}{oliver.lloyd@bristol.ac.uk}. \textbf{Supplementary information:} Supplementary data, including precision-recall curves and F1 curves for the best performing model, are available at \textit{Bioinformatics} online.
\end{abstract}

\keywords{polypharmacy side effect prediction \and adverse drug reaction \and knowledge graph embedding \and tensor factorisation}

\section{Introduction}

The comforts of modern life are causing a demographic shift in populations globally. By 2050, the number of people aged 60+ is projected to more than double \citep{AgeingWHO}, compared to an overall population increase of around 21\% \citep{PopulationUN}. With this increase, medical events that primarily affect older people will become increasingly important areas of research. Multimorbidity is one such phenomenon, whose prevalence among elderly populations may range from 55-98\% \citep{Marengoni2011}. Closely associated with multimorbidity is another issue: polypharmacy, the taking of two or more medications simultaneously by the same individual. Polypharmacy can be usefully employed in some contexts, for example in order to achieve drug synergism \citep{Tallarida2011}, where two medications are combined to produce an effect greater than the sum of their individual effects. However, the practise of polypharmacy can also lead to the emergence, via chemical interactions, of adverse drug reactions (ADRs) that are not associated with either drug individually \citep{Ahmed2014}. ADRs put a large strain on healthcare systems, with a systematic review from 2002 putting their annual cost to the UK National Health Service at £380 million \citep{Wiffen2002}. Given the general population increase and demographic shift mentioned above, plus inflation, it is very plausible that this annual cost may soon approach £1 billion.

Correct prediction of ADRs related to polypharmacy is a challenging task, owing mainly to the extremely large numbers involved. In England, an estimated 8.4 million people are taking at least 5 prescribed medications, and a quarter of those are taking 10 or more \citep{PrescribingGOV}. The combinatorial nature of the problem means that testing every possible n-combination of the thousands of commercially available drugs quickly becomes infeasible in wet-lab experiments or clinical trials, even with smaller values of n than are commonly found in real populations. As a result, pre-clinical screening with statistical/computational methods is a necessity for identifying drug combinations of interest \citep{Ryall2015}. Recently, there has been an explosion of interest in graph-embedding methods for solving the problem of polypharmacy side effect (PSE) prediction. This can largely be traced back to the work of \citeauthor{Zitnik2018}, in which the authors construct a knowledge graph from the following data: drug-target, protein-protein interaction, monopharmacy side effects, and drug-pair side effects \citep{Zitnik2018}. A portion of the drug-pair side effects are left out to enable out-of-sample prediction and assessment. By constructing the data in this way, the problem is cast as a multirelational link prediction (LP) problem, which, the authors claim, makes theirs the first technique that allows prediction of the \textit{type} of side effect that will occur, rather than a simple binary categorisation or magnitude of effect. The Decagon model itself consists of two components. Firstly, node embeddings for the network are encoded using a graph convolutional model. These embeddings are then passed to a tensor factorisation (TF) decoder which produces a score for a particular side effect \(r\) between a drug pair (\(v_i\), \(v_j\)). Finally, that score is passed to a sigmoid function which outputs a probability that the given triple (\(v_i\), \(r\), \(v_j\)) is true.

Commendably, \citeauthor{Zitnik2018} made their data publicly available and since then it has been used as a common benchmark in PSE modelling research. Early adopters of the dataset found success with a wide variety of techniques, including kernel ridge regression \citep{Dewulf2021}, semantic predication embedding \citep{Burkhardt2019}, and product of experts models \citep{Malone2019}. Each of these approaches improved upon the reported scores of Decagon while also being more efficient and easier to interpret. Despite this, as time has progressed we have seen an increasingly homogenised set of methods being applied to the Decagon dataset, with graph neural networks (GNNs) becoming dominant from 2022 onwards. GNNs have been reported to perform well on PSE prediction \citep{Carletti2021, Zhuang2023, Saifuddin2023}, particularly when information pertaining to the chemical structure of drugs is incorporated into the training process \citep{Li2023}. Unfortunately, there is no avoiding the fact that they are expensive to run \citep{Wu2019}, and this problem is set to only get worse for drug-structural models given that larger biologic drugs are becoming more prevalent over time \citep{Senior2023}.

TF models, a subset of the broader Knowledge Graph Embedding (KGE) family, are a set of techniques that embed KGs by formulating the graph as a third-order tensor before decomposing it into lower rank representations of the constituent entities and relations. They have proved useful in several real-world tasks, such as item recommendation \citep{Zhao2020}, drug development and repurposing \citep{Kim2023}, and knowledge base completion \citep{Kazemi2018, Trouillon2016}. The application of these techniques to the PSE modelling problem, however, has been somewhat scarce considering their relevance for this type of data alongside their relatively low cost to run and optimise. This is even more surprising in the light of research which has suggested that simpler models can in fact outperform more complex ones when correctly tuned \citep{Ali2022}. One possible explanation for this scarcity can be found in the original work of \citeauthor{Zitnik2018} - they employed two such methods (RESCAL and DEDICOM) alongside their graph convolutional approach, reporting that these two scored the lowest out of all the methods tested \citep{Zitnik2018}. Even predictions based on the simple method of concatenating drug features achieved substantially higher scores as measured by the three test metrics. In the years since that publication, there have been conflicting reports of the success of TF models. Papers have reported results close to those in the Decagon paper \citep{Wang2022b} through to the much more performant end of the scale \citep{Novacek2020, Dai2021}, with several more reporting scores somewhere in between \citep{Malone2019, Sserwadda2022,Li2023}.

Training speed is an important aspect of PSE prediction that is often overlooked and rarely reported on. The main benefit of faster models is that they lower the barrier to real-world implementation. Genetic differences between populations will change the structure of the protein-protein interaction subgraph, a core component of the input, such that a single global model would likely be insufficient in many cases. On the other hand, training models like NNPS \citep{Masumshah2021} or MDF-SA-DDI \citep{Lin2021} on a plethora of population-specific input datasets would incur substantial costs that could render this healthcare solution inaccessible to poorer communities. After all, cloud compute time is expensive, and hardware even more so. Shallower architectures that execute faster, such as DPSP \citep{Masumshah2023}, are more suited for the task but they still leave room for improvement. Further, a large portion of PSE research sensibly focuses on the inductive scenario (e.g. \citep{Zhu2024, Lin2023, Zhang2023, Li2023, Lukashina2022}) because new drugs are being developed frequently and a good model must be able to account for them. However, if we are able to reduce the cost of training down to almost-trivial levels, then the inductive scenario is made somewhat redundant because we can simply update the input data and re-train the model in a now-transductive context. This may be less useful to drug companies performing pre-clinical screenings, but in the patient-focused case where we are predicting effects between drugs that are currently on the market, there should be more than enough adjacency data to work with. Finally, papers that present cheaper models are also much easier to reproduce. Better reproducibility could lead to higher trust in the models being presented in the field, faster scientific progression through collaboration, and improved accessibility for newer researchers.

The contributions of this work are as follows. Firstly, we aim to contribute to the body of evidence on the predictive capabilities of TF methods on the PSE task. We do this by running a comprehensive hyperparameter optimisation procedure for each tested model within the LibKGE framework. This optimisation process involves Sobol sampling of the hyperparameter space followed by Bayesian modelling of embedding trial outcome against hyperparameter configuration. This is the first piece of PSE work to perform such a rigorous selection of hyperparameter values, and it allows us clear insight into the upper predictive limits of TF models on this task. As a secondary goal, we measure the change in predictive performance that occurs when incorporating monopharmacy data as reflexive edges in the graph, rather than using it to initialise embeddings as has been done previously \citep{Zitnik2018}. As far as we know, we are the first to model the monopharmacy data in this specific way, so we quantify and report the difference in performance that it causes. We round out the paper with a per-epoch analysis of our best performing model to see how its performance improves as a function of training progression.

\section{Methods}

\begin{figure}
\centering
    \textbf{A) Selfloops}
    \begin{subfigure}
        \centering
        
        \includegraphics[scale=0.75]{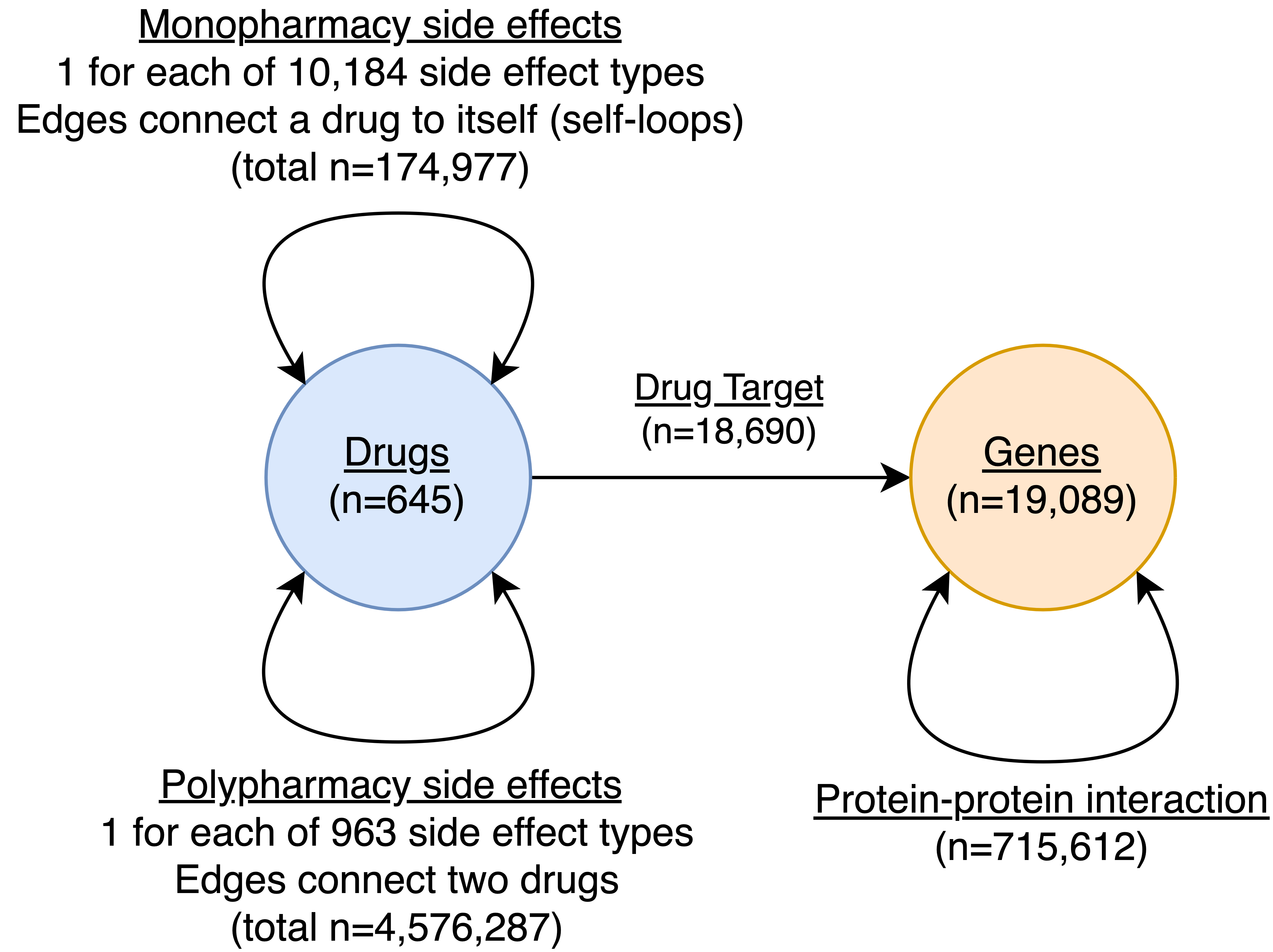}
        
        \label{non-naive}
    \end{subfigure}
    \vspace{10mm}
    \textbf{B) Non-naive}
    \vspace{5mm}
    \begin{subfigure}
        \centering
        
        \includegraphics[scale=0.75]{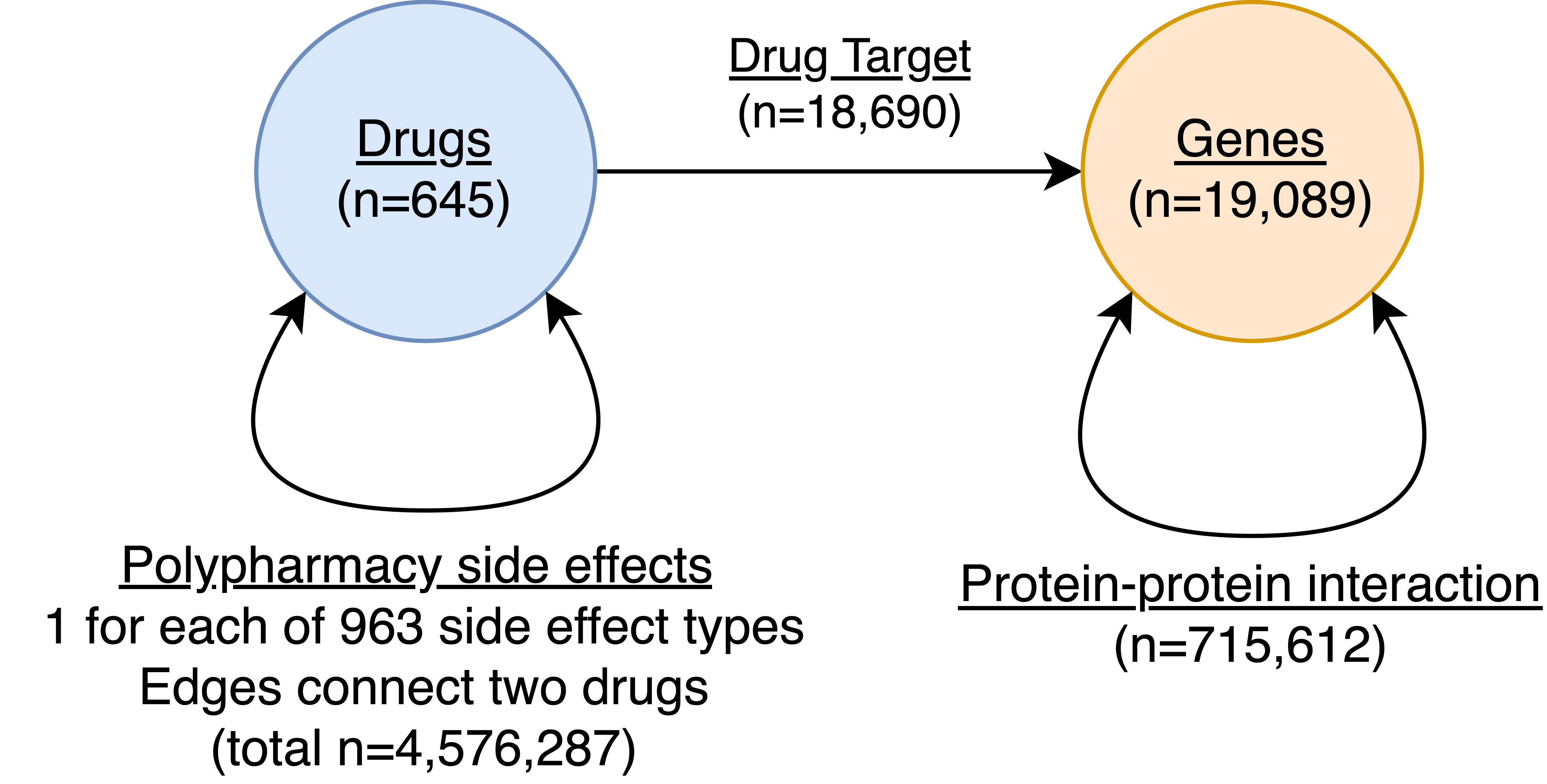}
        \label{Selfloops}
    \end{subfigure}
    \caption{Schemata for the two graph variants. The only difference between the two graphs relates to the handling of monopharmacy side effect data. The \textbf{Selfloops} approach (subfigure A) includes monopharmacy data as self-looping edges from a given drug back to itself, whereas the \textbf{Non-naive} graph (subfigure B) doesn't contain these edges. Instead, monopharmacy associations are modelled as n-hot vectors of length 10,184. At the start of an embedding trial, these vectors are reduced to the selected embedding sized using singular value decomposition and then used as the initialisation points for the drug embeddings. When training on the \textbf{Selfloops} graph, embeddings are initialised by drawing from a random distribution, as is standard in LibKGE.}
    \label{fig:schemata}
\end{figure}

We downloaded the raw data used by Decagon from the Stanford Network Analysis Project \nolinkurl{http://snap.stanford.edu/decagon/}. We then prepared the data for LibKGE \citep{Broscheit2020}, which reads graphs as a list of edges in RDF triple (subject, predicate, object) format. There are several ways to approach the conversion of the data to RDF format. In this work we test two such methods, which we have named ‘Selfloops’ and ‘Non-naive’ – the difference between them being the way in which monopharmacy side effect data is included in the resulting graph. The \textbf{Selfloops} approach treats these side effects as edges from one drug back to itself (hence the name). The \textbf{Non-naive} construction method is equivalent, but, following the example of \citeauthor{Zitnik2018}, monopharmacy data is instead modelled as n-hot node feature-vectors, with ‘hot’ columns indicating which side effects are associated with a given drug. We performed dimensionality reduction via principal component analysis (PCA) on these features to create a smaller matrix for each possible dimensional size of embeddings. We then modified LibKGE to load these vectors from disk, using them as the starting point for learning node embeddings for this dataset rather than any of its usual stochastic initialisation techniques. Graph statistics for the two networks are listed in table \ref{tbl:graphStats}, and schemata are shown in figure \ref{fig:schemata}.

\begin{table}
\begin{center}
\begin{tabular}{|c c c c c|}
 \hline
Graph & Meta-nodes & Nodes & Meta-edges & edges \\ [0.5ex]
 \Xhline{3\arrayrulewidth}
Selfloops & 2 & 19734 & 11149 & 5485566 \\
 \hline
Non-naive & 2 & 19734 & 964  & 5310589 \\
\hline
\end{tabular}
\newline
\caption{Node and edge counts for our two constructed variations of the Decagon graph. Meta-edges are the ‘types’ of edges allowed in the graph, such that each meta-edge can be associated with one adjacency matrix. A meta-node is a ‘type` of node that all follow the same connectivity rules and appear together as one or both axes of an adjacency matrix. For example, an instance of the ‘Drug-Target’ meta-edge can only exist between one ‘Drug’ node and one ‘Gene’ node, and can be represented as a \(d\) x \(g\) adjacency matrix in which a value of 1 indicates that drug \(d_i\) targets gene \(g_j\), otherwise the value is 0.}
\label{tbl:graphStats}
\end{center}
\end{table}

All three KGE methods included in this analysis can be classified as some form of TF. ComplEx \citep{Trouillon2016} uses the Hermitian product to embed entities into the complex space. This (ironically) straightforward approach allows anti-symmetric relations to be modelled with a low-rank decomposition, which, in the real space, would only be possible for symmetric relations. DistMult \citep{Yang2014}, also referred to as ‘bilinear-diag’ by its authors, is similar to the well known TransE model, but uses a multiplicative operation rather than an additive one to combine dyadic vectors. Lastly, SimplE \citep{Kazemi2018} is an enhancement of the Canonical Polyadic (CP) decomposition for embedding KGs. CP itself was used often in early LP research but has a major shortfall in that it learns head and tail vectors for a given node independently – SimplE addresses this problem by also considering inverted relations to create dependency between these vectors.

The hyperparameter optimisation space, established in configuration files, was the same for all models. Hyperparameter ranges were based on those from the search performed by the LibKGE developers in their demonstrative paper \citep{Ruffinelli2020}. We made a few adjustments to the space, notably expanding the number of available options for optimiser and loss function from 2 and 3, to 5 and 4 respectively. Other alterations included adding possible embedding sizes of 32 and 64, and reducing the aggressiveness of the learning rate scheduler. The grid searches took place over 100 trials, with the first 50 having values chosen by Sobol sequence and the remaining 50 chosen by Bayesian optimisation through the Ax platform (https://ax.dev). This Bayesian method works by iteratively fitting surrogate models (Guassian processes) of trial outcomes against the attempted hyperparameter configurations – these models are used to determine the ‘value’ of choosing a particular new parameterisation by considering both the predicted outcome of that configuration, as well as the corresponding uncertainty. The parameterisation that provides the best value is then chosen for the next trial, and afterwards the process is repeated. 

\begin{figure*}
    \centering
    \includegraphics[scale=0.75]{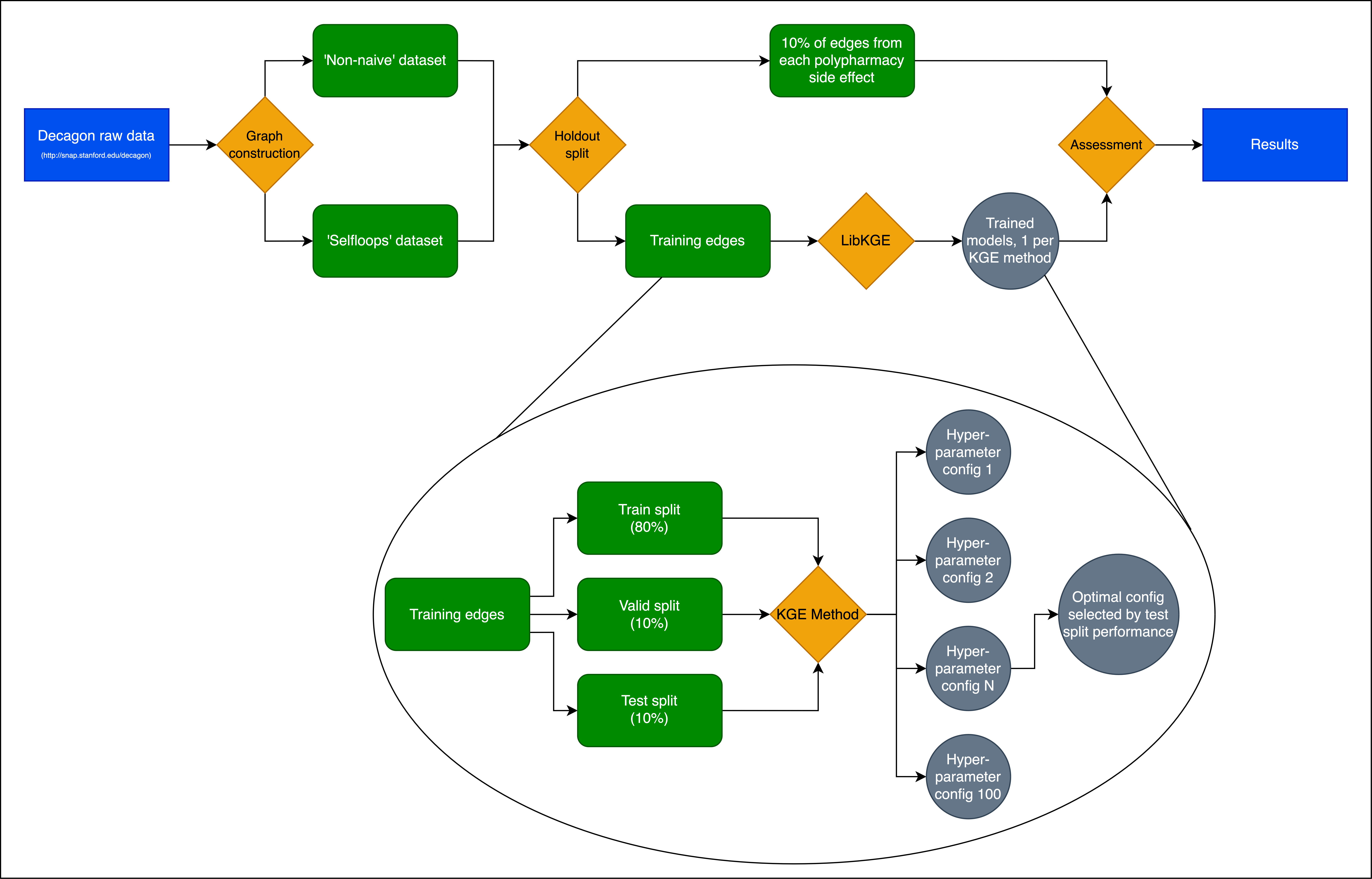}
    \caption{Workflow for the reported experiment. The datasets that are created from the raw data differ only in their handling of monopharmacy data. The holdout split is the same for both because they contain the same polypharmacy data. Both are then put through the workflow on the right hand side of the ‘holdout split’ node, and results compared. The lower diagram offers a ‘zoomed in’ look at the learning process in LibKGE that is not displayed in the upper diagram. This learning process happens once for each of the 6 dataset-model combinations. Key: blue = raw data; yellow = software/code; green = processed data; grey = trained model.}
    \label{fig:workflow}
\end{figure*}

We removed a portion of both test graphs, prior to the main analysis, for holdout validation. Following the methodology used by Decagon \citep{Zitnik2018}, the holdout data was created by randomly removing 10\% of the edges belonging to each PSE. Since the two test graphs incorporate the same polypharmacy data, the same edges were held out from both in order to enable fair comparison between the two datasets. This holdout set contains 458,061 edges, leaving a total of 5,761,807 edges in the \textbf{Selfloops} graph, and 5,586,830 in the \textbf{Non-naive} graph. During training, performance was measured using the standard LP metrics employed in LibKGE, namely mean reciprocal rank (MRR) and hits@k. Performance on the hold-out edges, our measure for overall quality, was assessed with the same metrics used by the Decagon authors - area under receiver-operating characteristic curve (AUROC), area under precision-recall curve (AUPRC), and average precision at 50 (AP@50). These were calculated individually per side effect type. A flowchart of the methodology of this paper is displayed in figure \ref{fig:workflow}.

The last piece of analysis involved identification of the best performing method/hyperparameter/dataset combination from the aforementioned testing. We re-ran the embedding procedure for this model, setting LibKGE to save the model state after every single epoch. After the embedding was complete, each of these model states was put through the same PSE testing procedure as the `final' model states were in the earlier section of this study. By assessing the model states in this way, we gain insight into the relationship between predictive performance and training progression. 

This work was carried out using the computational facilities of the Advanced Computing Research Centre, University of Bristol \nolinkurl{http://www.bristol.ac.uk/acrc}. The specific environment was CentOS-7 running Python 3.8.12 with PyTorch 1.7.1, accelerated with CUDA 11.4 on 4× NVIDIA GeForce RTX 2080 Ti. All code used in the experiments is available in our GitHub repository \nolinkurl{https://doi.org/10.5281/zenodo.10684402}.

\section{Results}
\subsection{Overall Performance}

\begin{figure}
    \centering
    \begin{subfigure}
        \centering
        \includegraphics[width=0.6\linewidth]{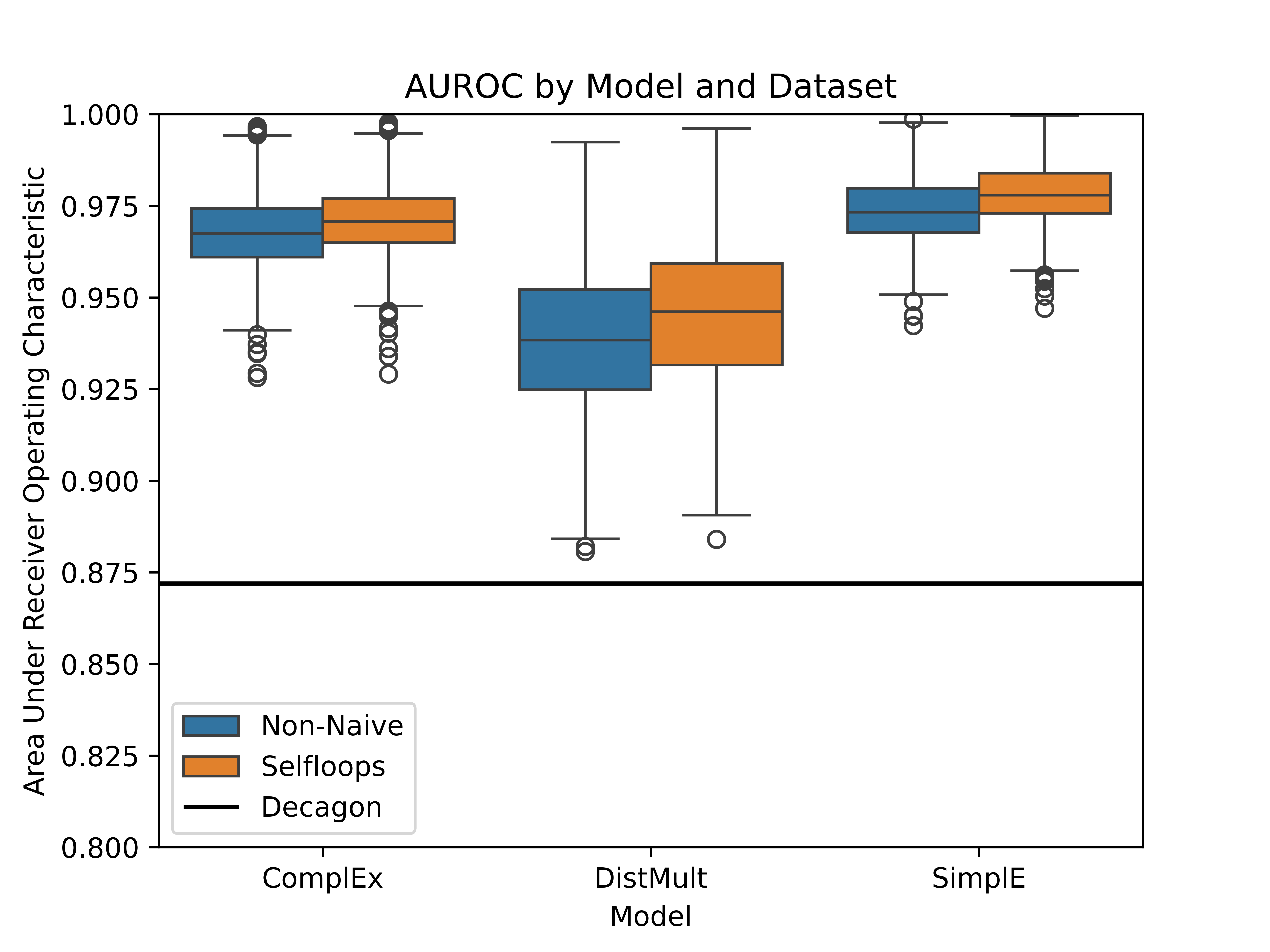}
    \end{subfigure}
    \begin{subfigure}
        \centering
        \includegraphics[width=0.6\linewidth]{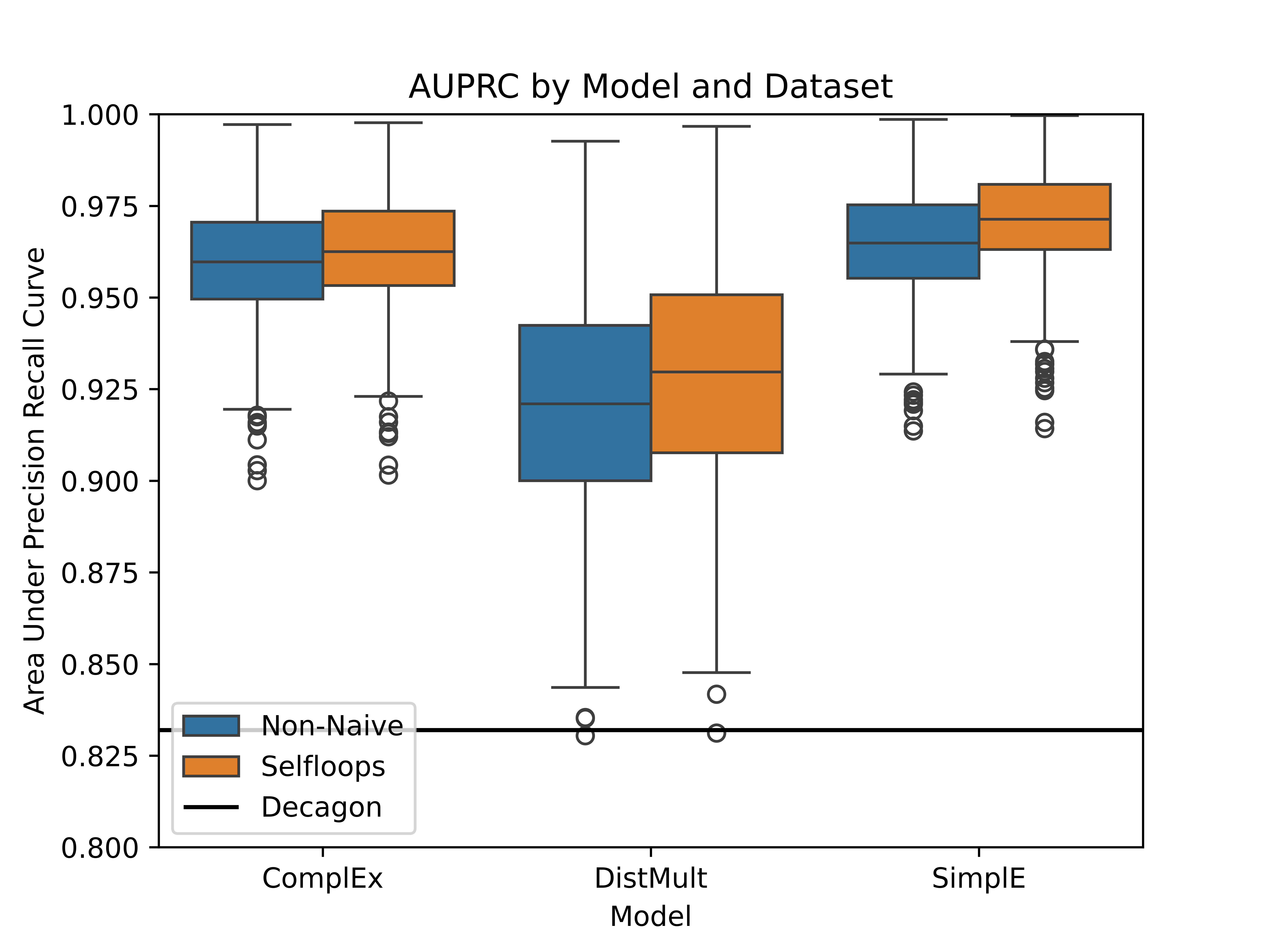}
    \end{subfigure}
    \begin{subfigure}
        \centering
        \includegraphics[width=0.6\linewidth]{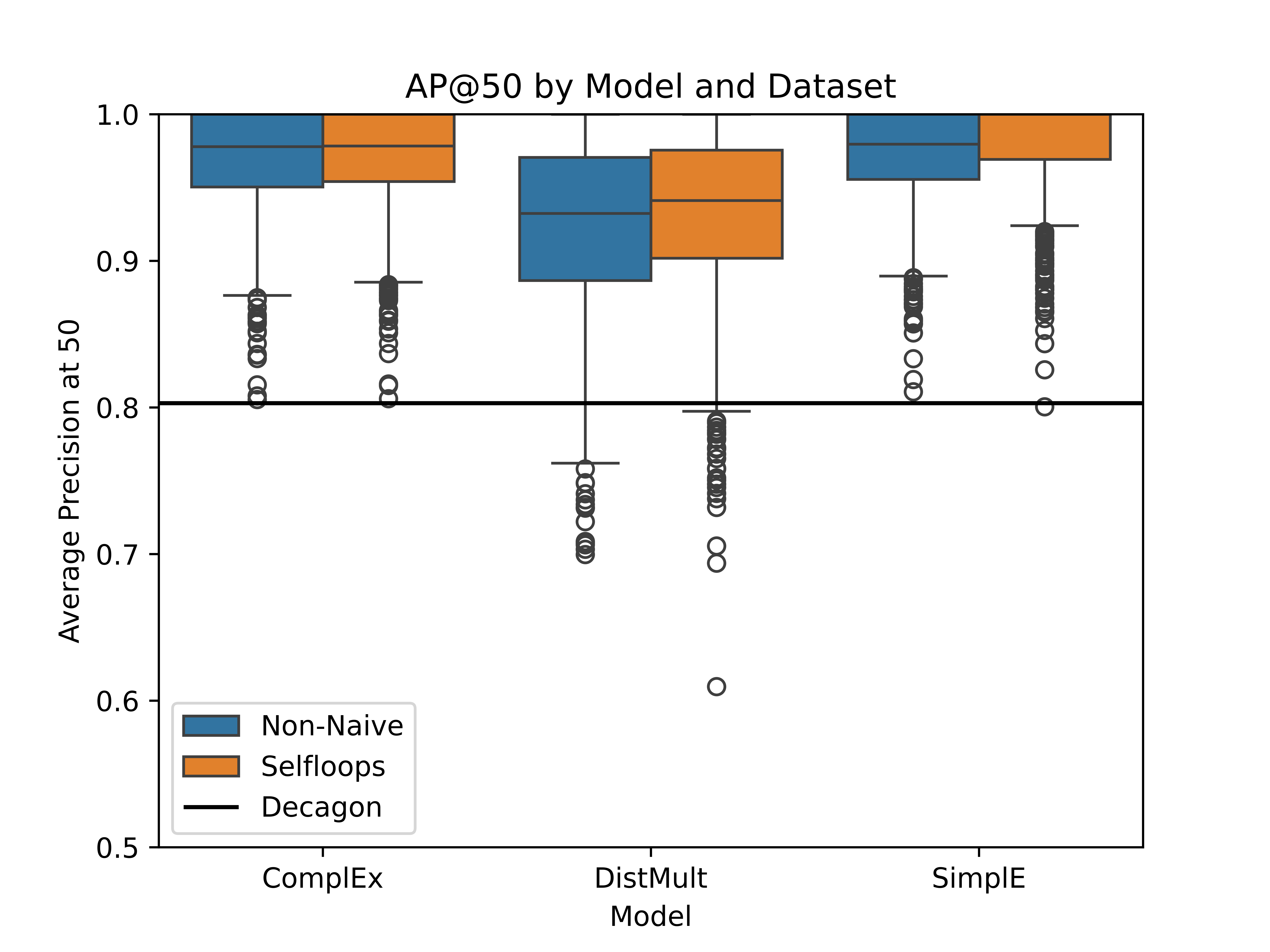}
    \end{subfigure}
    \caption{Out-of-sample prediction performance over 963 side effects as measured by three metrics. Top: Area Under Receiver Operating Characteristic; Middle: Area Under Precision Recall Curve; Bottom: Average Precision at 50. Results are presented for each of 6 experiments, comprising 3 models (ComplEx, DistMult, SimplE) run on two datasets (\textbf{Non-naive}, \textbf{Selfloops}). Decagon's performance on the same data is indicated by the horizontal line.}
    \label{fig:performance}
\end{figure}

Results of the assessment are shown in figure \ref{fig:performance}. All three of our tested KGE methods outperformed Decagon on both of our constructed graphs, with median scores improving on Decagon's performance by 7.62 - 12.2\% for AUROC, 10.7 - 16.8\% for AUPRC, and 16.1 - 24.5\% for AP@50. Remarkably, the rankings of our six configurations by median score do not change regardless of which metric is used to order them (table \ref{tbl:perfSummary}). Full precision-recall curves, as well as the corresponding F1 curves, of our best model (\textbf{SimplE-Selfloops}) are contained in the supplementary material of the paper.

\begin{table}
\begin{center}
\begin{tabular}{|c c c c c|} 
 \hline
 Model & Dataset & AUROC & AUPRC & AP@50 \\ [0.5ex] 
 \Xhline{3\arrayrulewidth}
 SimplE & Selfloops & 0.978 & 0.971  & 1.000 \\ 
 \hline
 SimplE & Non-naive & 0.973 & 0.965  & 0.980 \\ 
 \hline
 ComplEx & Selfloops & 0.970 & 0.963  & 0.978 \\ 
 \hline
 ComplEx & Non-naive & 0.967 & 0.960  & 0.978 \\ 
 \hline
 DistMult & Selfloops & 0.946 & 0.930  & 0.941 \\ 
 \hline
 DistMult & Non-naive & 0.938 & 0.921  & 0.932 \\ 
 \hline
\end{tabular}
\newline
\caption{Median performance across 963 polypharmacy side effect types as measured by three metrics for each of our six experiments.}
\label{tbl:perfSummary}
\end{center}
\end{table}

To provide some context to the performance of our best model, SimplE on the \textbf{Selfloops} graph, we gathered information about the highest performing methods that have been presented in the literature. The top 10 of these, as ranked by AUPRC, are shown in table \ref{tbl:otherModelPerformance}.

\begin{table*}
\begin{center}
    \begin{tabular}{|c c c c c c|} 
     \hline
     Model & AUPRC & Mechanism & Complexity & Extra data & Citation \\  
     \Xhline{3.5\arrayrulewidth}
     Carletti's & 0.998 & GAT & $O(d^2)$ & None & \citep{Carletti2021} \\ 
     \hline
     MS-ADR & 0.983 & GCN & $O(d^2)$ & Enzyme and transporter & \citep{Zhuang2022} \\ 
     \hline
     ADGCL & 0.980 & GNN & $O(d^2)$ & Enzyme and transporter  & \citep{Zhuang2023} \\ 
     \hline
     EmerGNN & 0.976 & GAT & $O(d^2)$ & Hetionet & \citep{Zhang2023} \\
     \hline
     GS-ADR & 0.972 & GAT & $O(d^2)$ & Enzyme and transporter  & \citep{Zhuang2021} \\ 
     \hline
     SimplE & 0.971 & TF & $O(d)$ & None & (This work) \\ 
     \hline
     SimVec & 0.968 & Auto-Encoder & $O(d)$ & Substructure  & \citep{Lukashina2022} \\ 
     \hline
     HyGNN & 0.965 & GCN & $O(d^2)$ & Substructure  & \citep{Saifuddin2023} \\ 
     \hline
     HLP & 0.965 & GNN & $O(d)$ & Substructure  & \citep{Vaida2019} \\ 
     \hline
    
     DeepDrug & 0.960 & GNN & $O(d^2)$ & Substructure  & \citep{Yin2023} \\ 
     \hline
    \end{tabular}

\caption{Top 10 performing models by AUPRC on the Decagon dataset. GAT = Graph Attention Network, GCN = Graph Convolutional Network, GNN = Graph Neural Network, TF = Tensor Factorisation. Model complexity is given in big O notation as it relates to the embedding dimensionality (d).}
\label{tbl:otherModelPerformance}
\end{center}
\end{table*}

On our investigation of the mechanism for including monopharmacy data into the embedding pipeline, we find that the \textbf{Selfloops} graph achieves better median results than the \textbf{Non-naive} graph for all three embedding methods. In all cases, however, this improvement is \(\leq\) 0.02, and in the case of the ComplEx model measured by AP@50, the difference is as low as 0.0004. 

We observed varying hyperparameter configurations of the optimal models from each experiment. Either 1vsAll or KvsAll was the chosen method in all cases for the sampling of negative training examples, with the top (\textbf{SimplE-Selfloops}) and bottom two (DistMult) models using 1vsAll and the others using KvsAll. The chosen loss functions matched this pattern exactly — BCE was used under the KvsAll strategy, and Kullback–Leibler divergence (KL) used under 1vsAll. Adamax was the modal optimiser, chosen in four out of six cases. The two exceptions, \textbf{SimplE-Selfloops} and \textbf{ComplEx-Selfloops}, used Adam and Adadelta, respectively. Dimensionality also varied between the embeddings, with half the models opting for m = 128, two using m = 256, and just one employing m = 512. Since \textbf{Non-naive} initialises embedding values with principal components of the n-hot monopharmacy side effect vector, the ‘weight initialisation’ hyperparameter was only relevant for the models applied to the \textbf{Selfloops} dataset. Within these, ComplEx and DistMult simply drew values from a normal distribution, whereas SimplE subsequently applied the ‘Xavier’ \citep{Glorot2010} modification to these values.

\subsection{Training Speed}
\begin{figure}
    \centering
    \begin{subfigure}
        \centering
        \includegraphics[width=0.7\linewidth]{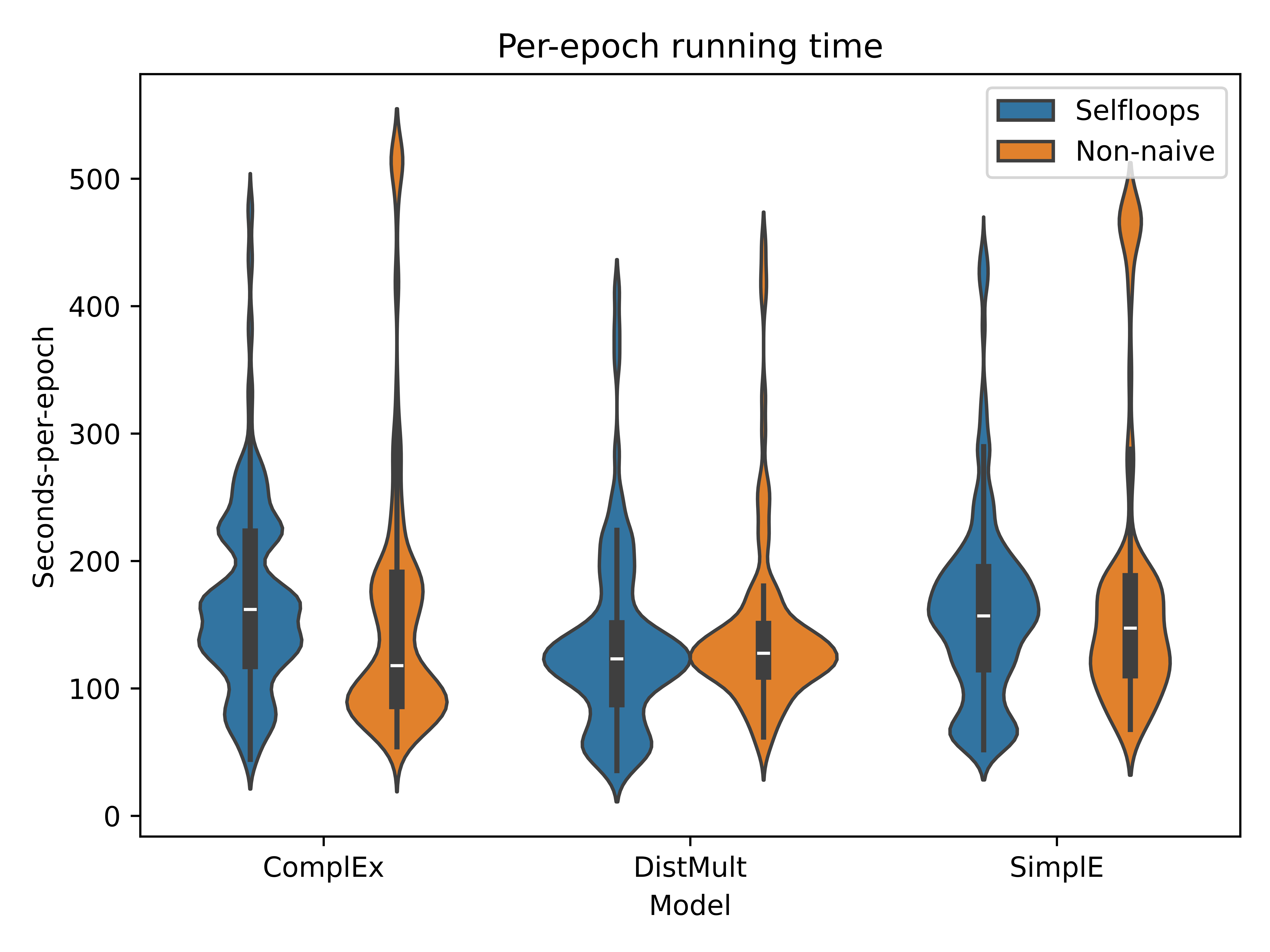}
    \end{subfigure}
    \begin{subfigure}
        \centering
        \includegraphics[width=0.7\linewidth]{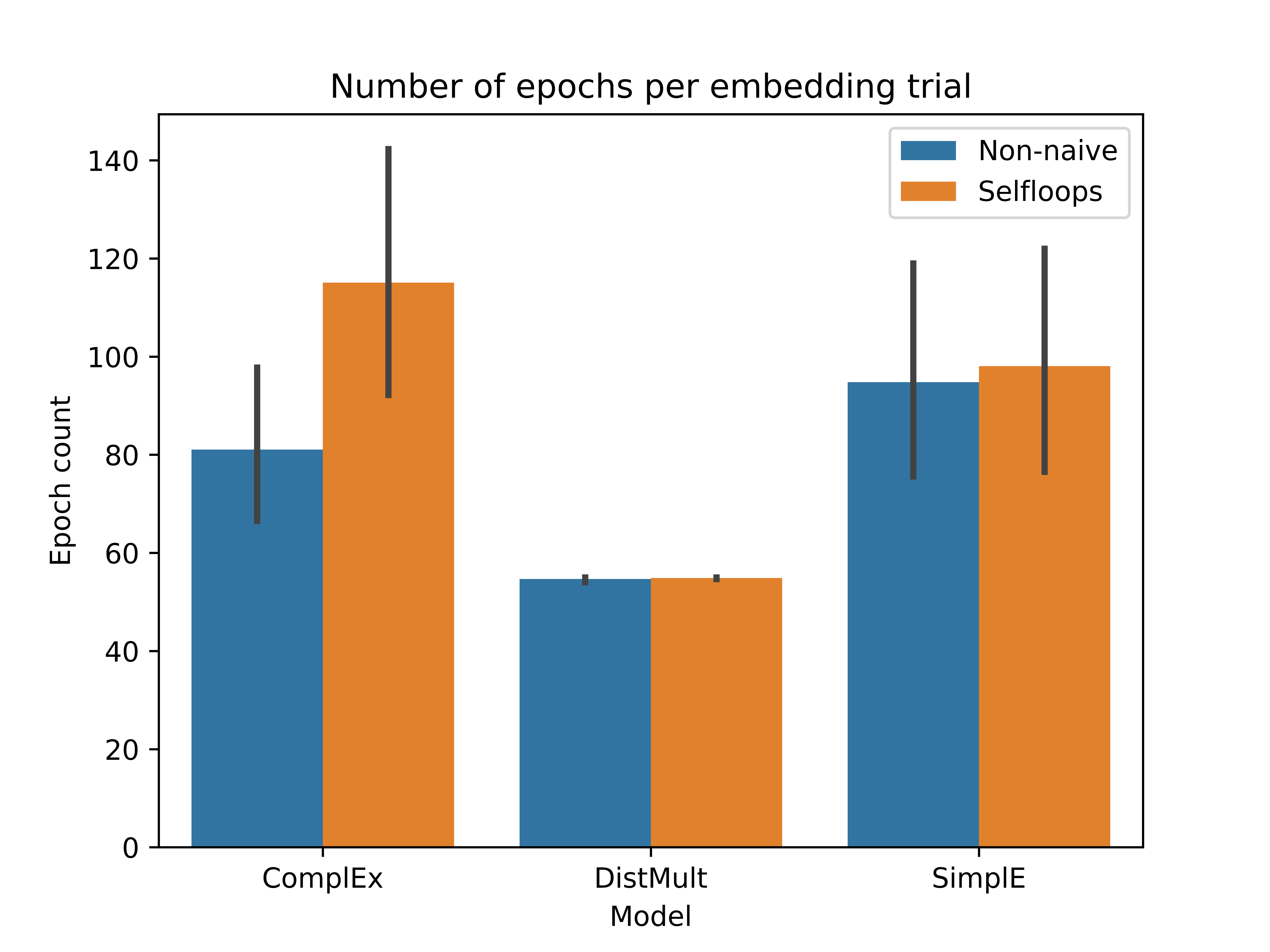}
    \end{subfigure}
    \caption{Running time information for the six experiments. Subfigure A is a violin plot showing the distribution of compute time per-epoch. Subfigure B shows a bar plot of the mean number of epochs per-trial, with a black bar representing the standard error. An early stopping procedure was used to determine the maximum number of epochs in a trial, with a minimum of 55 and a maximum of 500.}
    \label{fig:runtime}
\end{figure}

Each trial, of the 100 per experiment, ran for a variable number of epochs that was determined by an early stopping procedure. This procedure checks the performance of the trial every five epochs starting from 50, ending the trial if the performance has not improved after two consecutive checks. Consequently, to compare the speed of computation between experiments we first standardized the time by dividing it by the total number of epochs run. Figure \ref{fig:runtime}.A shows these values. Although DistMult is the fastest, with a mean time-per-epoch of 165 seconds on \textbf{Selfloops} and 183 on \textbf{Non-naive}, we observe surprising consistency overall with only a small gap from DistMult to the other models. The slowest overall was SimplE on \textbf{Non-naive}, achieving an average epoch completion time of 219 seconds. Figure \ref{fig:runtime}.B shows the mean number of epochs (plus standard error) that the trials ran for in each experiment. The DistMult and SimplE models had comparable epochs-per-trial between the two datasets, with the former always stopping at the minimum count of 55. Interestingly, ComplEx was about 30\% faster on \textbf{Non-naive}, with a mean epoch time of 81 seconds versus 115 for the slightly larger graph \textbf{Selfloops}. 

\begin{figure}
    \centering
    \begin{subfigure}
        A)
        \centering
        \includegraphics[width=0.7\linewidth]{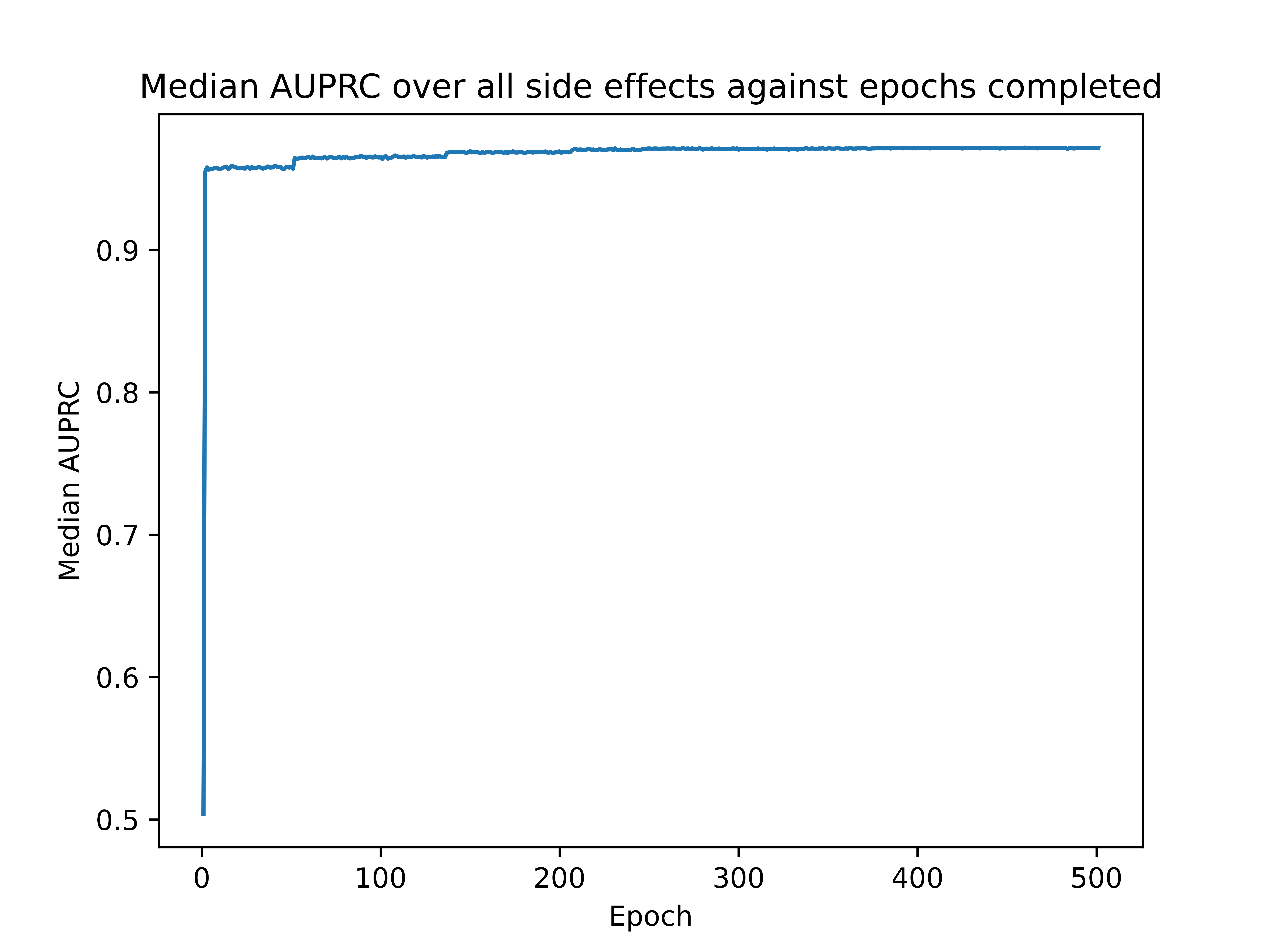}
    \end{subfigure}
    \begin{subfigure}
        B)
        \centering
        \includegraphics[width=0.7\linewidth]{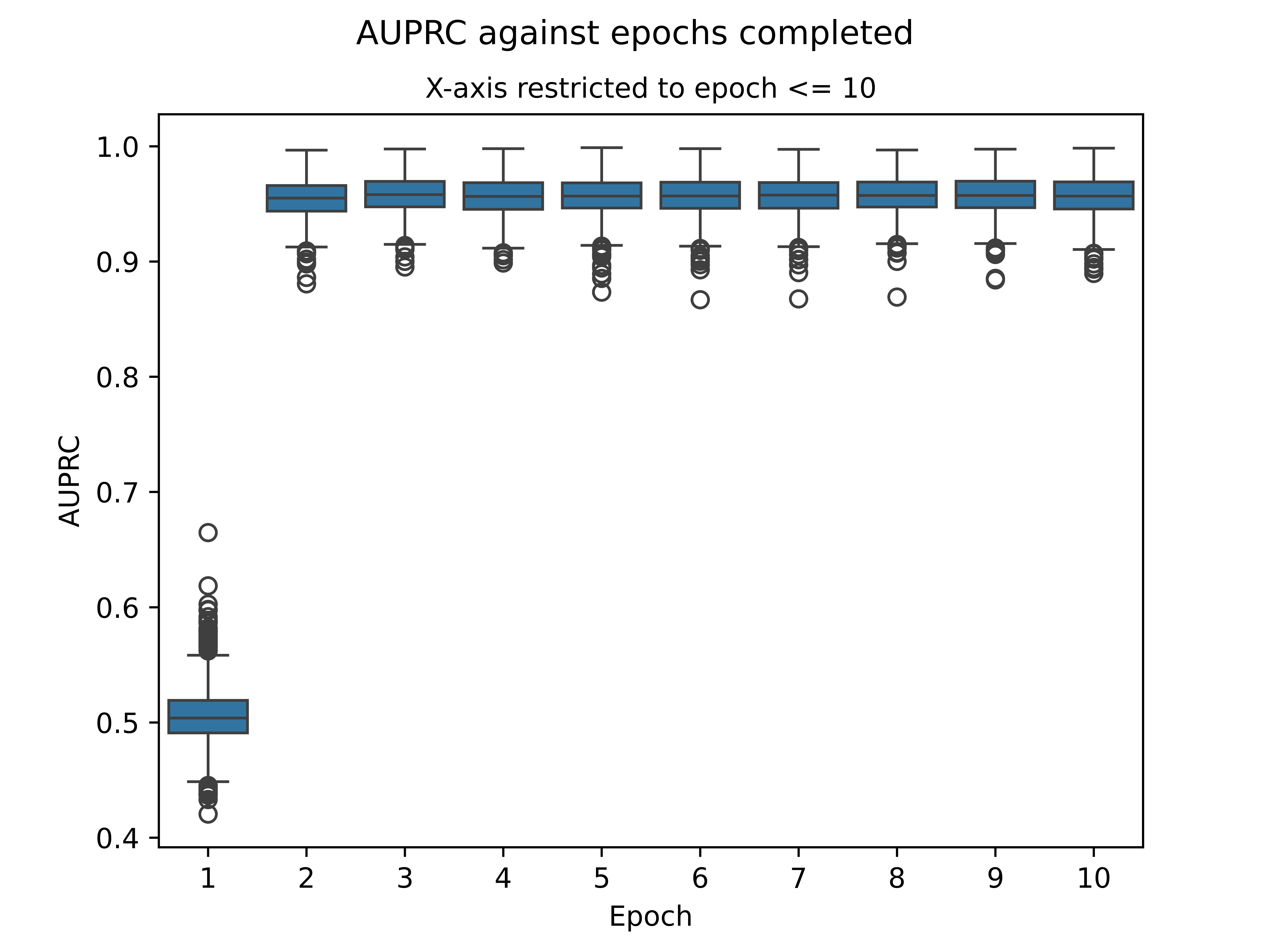}
    \end{subfigure}
    \caption{Side effect prediction performance after every epoch of training, for our best performing model. Subfigure A shows the median performance (across the 963 side effects) for all training epochs. Subfigure B offers a closer look at the first ten epochs, for each of them showing the distribution of the models performance over all side effects as a box plot.}
    \label{fig:per_epoch}
\end{figure}

Figure \ref{fig:per_epoch} shows the performance-per-epoch of our best performing model - SimplE on the \textbf{Selfloops} graph variant. After just two epochs, which took a total of 214.7 seconds, the model is already predicting side effects with a median AUPRC of 0.955. The highest overall median AUPRC was 0.972, achieved after epoch 460 (49,543 seconds), meaning that the model reached 98.3\% of its best performance after 0.43\% of the required training time. Equivalent relationships were observed for the other two metrics, AUROC and AP@50.

\section{Discussion}
The overall predictive performance reported here ranks among the best to date - in fact, there are only four models that claim to achieve greater AUPRC (table \ref{tbl:otherModelPerformance}). The highest of these four, \citeauthor{Carletti2021}'s model, extends Decagon's architecture by adding a Relation Attention Module (RAM) which "increases the expressiveness of the network by weighting the contribution of the messages exchanged over different relations". This relatively simple modification to the Decagon method reportedly causes a drastic improvement in prediction performance, raising both AUROC and AUPRC to 0.998 from 0.872 and 0.832 respectively. However, the authors provide no publicly available repository to enable reproduction of their work and, whilst they did send archived code on request, the lack of documentation meant we were unable to run it. The absence of the original coder from their team also meant that they could provide no technical assistance and as a result we were unable to reproduce their reported model performance. The 2nd to 4th ranked methods all come from the same pair of authors, Luhe Zhuang and Hong Wang of Shandong Normal University. All three of their approaches use the same dataset - a modified version of Decagon's graph, with enzyme and transporter data added and 97 drug nodes removed. It is unclear which drug nodes were removed and no motivation for doing so is given. The earliest presented of their models, GS-ADR, focuses on the concept of signed edges in the graph, with the paper reporting that consideration of this results in ‘more effective’ feature representation of drugs \citep{Zhuang2021}. MS-ADR, published in 2022, also uses signed edges but additionally considers four different biomedical ‘views’ (enzyme, indication, side effect, and transporter) when convolving information across the graph \citep{Zhuang2022}. ADGCL is the latest of their models and is similar to MS-ADR but makes use of an Implicit GNN rather than a convolutional encoder, finding that comparable performance is achieved \citep{Zhuang2023}. As with \citeauthor{Carletti2021}'s work, none of these three papers offer a publicly available repository of the code used to enable replication of their findings. Although AUPRC was not measured in their paper, it is well worth mentioning the work of \citeauthor{Li2023}, who report an AUROC of 0.999 on the Decagon dataset by incorporating drug atomic structure into the network \citep{Li2023}. In this case, drugs are viewed as both singular graphs, where edges are bonds between atomic nodes, and also bipartite graphs, where potentially interacting drug pairs have edges from each atom to all atoms in the other molecule. By incorporating both the ‘inter’ and ‘intra’ drug views into a co-attention decoder, \citeauthor{Li2023} achieve what is arguably the current state of the art when it comes to predictive performance on the Decagon data. 

Thanks to our per-epoch testing, we are able to look beyond the overall predictive results and report an important finding regarding training speed: the SimplE model \citep{Kazemi2018}, trained on a six-year-old GPU (Nvidia RTX 2080 Ti), is able to learn embeddings that are highly predictive of PSEs in \textit{less than four minutes} (figure \ref{fig:per_epoch}). This result has substantial implications for real-world applications of PSE models because it drastically improves scalability by lowering the cost of model training/re-training compared to other techniques, while maintaining a comparable level of predictive power. No other model presented in the literature comes close to this speed. EmerGNN \citep{Zhang2023} is likely the closest, reaching a similar AUPRC in just under an hour (see figure 2b of that paper), though that model did have the benefit of training on the more powerful RTX 3090 according to the project repository. Generally speaking, very few PSE papers report training curves at all, so it is difficult for us to specifically quantify the improvement that SimplE brings. What we can say is that these results offer a motivation for the field to begin considering more ``green AI'' approaches, when currently the predominant models fall firmly in the ``red AI'' camp \citep{Schwartz2020}. 

It is not immediately apparent \textit{why} SimplE trains so fast on this data. The most similar published work was written by \citeauthor{Novacek2020}, whose TriVec model (also implemented in LibKGE) reached a slightly higher level of performance than our two-epoch result after training for a fixed 1,000 epochs on the same data \citep{Novacek2020}. Comparisons between their experiment and \textbf{SimplE-Selfloops} are displayed in table \ref{tbl:TriVec_params}. The largest difference between the experiments lies in the `batch size' hyperparameter, which was set to 6,000 for TriVec compared to just 256 for SimplE. It is generally regarded that a lower batch size will increase speed of convergence \citep{Kandel2020}, but this factor alone seems insufficient to cause such a speed-up considering that: A) SimplE embedded to more than twice as many dimensions as TriVec; and B) SimplE started at a learning rate an order of magnitude lower than TriVec. The Adam optimiser might converge faster than Adagrad \citep{Kingma2015}, but we wouldn't expect to see much difference between them after just two epochs (i.e. before momentum and weight-decay have kicked in). As a consequence of all of this, and because no training curve was presented alongside TriVec, we cannot rule out the simplest explanation that this impressive speed is just a general feature of TF on the Decagon dataset. In other words, both ourselves and the TriVec authors may have used far more resources on model training than was actually necessary.

\begin{table*}
\begin{center}
\begin{tabular}{|c c c |} 
 \hline
 Hyperparameter & TriVec value & SimplE value \\  
 \Xhline{3\arrayrulewidth}
 Optimiser & Adagrad & Adam  \\ 
 \hline
 Learning rate & 0.1  & 0.011 * \\ 
 \hline
 Embedding size & 100 & 256  \\ 
 \hline
 Batch size & 6000 & 256  \\ 
 \hline
 Regularisation weight & 0.03 & 0 \\ 
 \hline
 Dropout & 0.2 & Entity vectors: 0.068; Relation vectors: 0.125  \\ 
 \hline
\end{tabular}
\newline
\caption{Experiment comparison between TriVec \citep{Novacek2020} and \textbf{SimplE-Selfloops}, the best performing model from this study. * Learning rate was variable in the overall experiment but started at 0.011 and remained at this value for at least the first two epochs, which are our epochs of interest.}
\label{tbl:TriVec_params}
\end{center}
\end{table*}

We do not suggest that the two-epoch result reported here means that research into expensive methods should stop altogether - there will always be space in the field for ``red'' models. For example, GNNs do have a definite advantage when it comes to inductive reasoning. TF methods, as they currently exist, simply cannot make inferences about unseen nodes and would require the entire graph to be re-embedded if predictions about new drugs were required. In the patient-focused scenario, where even newly prescribed drugs will have a good level of side effect adjacency data available, this is no problem because (re-)embedding can be done very fast, as we have shown. However, for pharmaceutical companies looking to predict adverse interactions for pre-market drug candidates, TF would be wholly insufficient. In this case, complex models that consider the intricate molecular structures of drugs would serve them far better. Another limitation of both GNNs and TF is the restriction to drug dyads, when, in real life, patients may often be co-prescribed three or more drugs \citep{PrescribingGOV}. Unique interactions can occur in such cases, e.g. when taking skeletal muscle relaxants \citep{Chen2022}, so a clinically useful ‘polypharmacy’ side effect prediction model might need to be able to consider more than just the ‘duopharmacy’ case. As such, this is another area where more complex models may have an advantage. Hypergraph embedding models, such as those used by HyGNN \citep{Saifuddin2023} and HLP \citep{Vaida2019}, seem the most promising candidates to solve this particular challenge because, by definition, hyperedges can connect any number of nodes. However, research in this direction has so far been scarce, probably due to the lack of available training data available for such a task.

In this work we observed a slight improvement in predictive performance when including monopharmacy information as self-looping edges in the graph, compared to using it to initialise embedding vectors. This improvement held for all metrics and models, with the exception of ComplEx where an AP@50 score of 0.978 was achieved on both graphs. The level of improvement was at best 0.02, however, so the self-looping form of the dataset should only be considered ‘better’ if the overall efficiency of the model is not noticeably hampered by the addition of the 174,977 extra edges. As shown in figure \ref{fig:runtime}.A, runtime-per-epoch was not notably different between the two datasets when using the DistMult or SimplE models. ComplEx did take 34 seconds longer on average (81 vs 115) when embedding Selfloops, so in this particular case it may actually be worthwhile to initialise embeddings using monopharmacy data, as was done by \citeauthor{Zitnik2018}. Generally though, we can conclude that the monopharmacy side effect data should be modelled as graph edges instead when working with TF models.

In summary, we have shown here that TF embedding methods can rival state-of-the-art GNNs on the task of PSE prediction, even when trained with less data. Further, we have shown that the SimplE model can be trained to almost-full power at lightning speeds. Given the nature of the challenge, scalability should be a primary concern for anyone implementing such solutions, and TF models like SimplE provide a huge advantage in this regard. They also bring the benefit of lower electricity consumption and therefore a reduced carbon footprint, which is, of course, an important consideration in the context of climate change. \citeauthor{Zitnik2018} reported very poor performance of their only tested TF models, with both RESCAL and DEDICOM being outperformed even by a simple predictor. This may well explain the relative lack of interest in TF by later researchers. However, when set up with optimal hyperparameters and run with modern loss functions and sampling strategies such as those implemented in LibKGE, we have demonstrated that they could be more useful than the current state of the art. 

\section{Competing interests}
T.R.G. receives funding from Biogen and GSK for unrelated research.

\section{Author contributions statement}

\textbf{Oliver Lloyd}: Conceptualization, Methodology, Software, Formal analysis, Investigation, Data Curation, Writing - Original Draft, Writing - Review \& Editing, Visualization.\newline
\textbf{Yi Liu}: Conceptualization, Writing - Review \& Editing, Supervision.\newline
\textbf{Tom R. Gaunt}: Conceptualization, Writing - Review \& Editing, Supervision, Project administration, Funding acquisition

\section{Acknowledgments}
The authors thank Patrick Rubin-Delanchy for his helpful input.

\bibliographystyle{abbrvnat}  
\bibliography{references}

\end{document}